\documentclass[10pt,twocolumn,letterpaper]{article}

\usepackage[pagenumbers]{wacv}       
\definecolor{linkblue}{rgb}{0.21,0.49,0.74}
\usepackage[breaklinks,colorlinks,allcolors=linkblue]{hyperref}

\title{Ten Architectures, One Error: Shared Failure Modes in Hyperspectral Classification under Spatially Disjoint Evaluation}

\author{
Ehsan Faghih \qquad Fatemeh Ashrafi \qquad Marguerite Moore \qquad Zahra Saki\\
North Carolina State University, Wilson College of Textiles\\
Raleigh, NC, USA\\
{\tt\small sfaghih@ncsu.edu \quad fashraf@ncsu.edu \quad marguerite\_moore@ncsu.edu \quad zsaki@ncsu.edu}
}

\begin{document}
\maketitle
\begin{abstract}
Hyperspectral image classification still relies heavily on random pixel splits within a single scene. The Salinas dataset, randomly split, is among the most widely used datasets for comparing different architectures. However, under a random split method, a large fraction of test pixels fall immediately adjacent to a training pixel, which inflates reported accuracy. This work introduces a leakage-free evaluation protocol linking spatial separation to the model's receptive field. Applying this protocol across ten different architectures, including classical, spectral, spectral-spatial, transformer, vision-backbone, and state-space families, shows that Macro-F1 drops by $0.147$ on average and model rankings change by as many as five places. Furthermore, leakage-free evaluation limits which architectures can be tested on a given benchmark. Since each partition supports patches only within a finite radius, reporting this radius alongside the receptive field is essential for fair comparison. In addition, this study reveals that all ten architectures misclassify largely the same pixels, pointing to a spectral ambiguity in the data that none of them resolves.
\end{abstract}

\section{Introduction}
\label{sec:intro}
 
Hyperspectral image (HSI) classification has historically been benchmarked on a small recurring set of single-scene datasets, particularly Indian Pines, Pavia University, and Salinas, with training and test pixels commonly sampled at random from within the same scene, yielding very high reported Overall Accuracy (OA) and performance metrics~\cite{nalepa2019validating, morales2021hyperspectral, liu2017svm, roy2020hybridsn, zhang2023lightweight, khan2024groupformer}. This evaluation practice remains common in recent work. In 2026, Huang \etal~\cite{huang2026sfcfnet}
randomly allocated only $0.5\%$ of the labeled Salinas pixels to training, $0.5\%$ to validation, and the remaining $99\%$ to testing and, nevertheless, reported $99.05\%$ OA. Similarly, Zhong \etal~\cite{zhong2026lctnet} used random $1\%$ / $1\%$ / $98\%$ (training/validation/test) partitions on Salinas and reported $99.31 \pm 0.26\%$ OA. These near-ceiling results persist despite extremely limited labeled training sets. Meanwhile, HSI classifiers have progressed from convolutional architectures~\cite{roy2020hybridsn} to
spectral transformers~\cite{hong2022spectralformer}, state-space models~\cite{wang2025s2mamba}, and foundation models~\cite{wang2025hypersigma}, often evaluated under heterogeneous dataset splits and training regimes.
 
Random pixel splitting is not neutral for spatially autocorrelated data. Adjacent pixels within homogeneous regions are strongly spatially correlated, so a test pixel drawn in close proximity to a training pixel provides less independent evidence of generalization~\cite{wang2023spatial}. Patch-based models compound the problem: during training, their receptive fields cross the partition boundary and expose pixels that belong to the neighboring test partition~\cite{nalepa2019validating}. The ecology and geostatistics literatures have measured this effect directly. Roberts \etal~\cite{roberts2017crossvalidation} showed that ignoring dependence structure underestimates predictive error, and proposed blocked cross-validation as the remedy. In a drone-based tree-species segmentation study, Kattenborn \etal~\cite{kattenborn2022spatially} found that random validation overestimated CNN performance by up to $28\%$. Similarly, when Ploton \etal~\cite{ploton2020spatial} applied spatial validation to tropical biomass models, the explanatory power reported under random validation fell to near zero. Blocking, however, is specified in terms of distance between partitions, not in terms of what a model actually consumes; for patch-based classifiers the two differ, and the gap is where leakage survives. Mainstream vision has confronted a different form of leakage: $3.3\%$ of CIFAR-10 and $10\%$ of CIFAR-100 test images duplicate a training image, and the benchmark was reissued once this was measured~\cite{barz2020cifair}. Within HSI, the concern has surfaced twice, and both studies released a leakage-aware split~\cite{nalepa2019validating,zou2020ss3fcn}.

Existing leakage-aware studies demonstrate the problem on limited sets of architectures~\cite{nalepa2019validating,zou2020ss3fcn}. Neither reports the spatial separation its partition achieves nor re-evaluates a cross-generational set of models under an equal hyperparameter budget. Whether the rankings this literature reports survive spatially disjoint evaluation is therefore untested. This work answers that question on Salinas. The first contribution is a leakage-free evaluation protocol whose admissibility criterion ties spatial separation to model receptive field, with a frozen hash-verified split and a released common evaluation subset on which every model is scored identically. The second is a reusable audit procedure, applicable to any benchmark built from spatially structured data, covering partition disjointness, receptive-field admissibility, mask-composition bias, and a spectral probe that measures the protocol's effect before any model is trained. The third contribution is a benchmark of ten architectures spanning classical, spectral, spectral-spatial, transformer, vision-backbone, and state-space families. Every model receives the same budget of 90 hyperparameter trials, convergence is checked and reported per model, and results are averaged over five seeds. Published comparisons in this literature rarely hold tuning effort equal across architectures, so reported rankings partly reflect how hard each model was tuned rather than the architectures themselves.

Leakage-free evaluation changes what the benchmark reports. Rankings shift by as many as five places: SpectralFormer~\cite{hong2022spectralformer} ranks third under random splitting but falls to eighth under spatially disjoint evaluation, behind a tuned RBF-SVM~\cite{melgani2004svm} and a random forest~\cite{belgiu2016random}. At matched training-set size, every model performs worse, with a mean drop of $0.147$ Macro-F1 and a maximum drop of $0.228$. The architectures differ substantially in Macro-F1, yet they fail on the same pixels. On a single seed, $83\%$ of the wrong predictions on universally misclassified pixels are one directional confusion between two classes that are never spatially adjacent in the scene; across all five seeds, $1{,}156$ pixels are misclassified by all ten models. The limitation is spectral, not spatial. Finally, the training partition supports patches only up to a radius smaller than HyperSIGMA's published configuration requires~\cite{wang2025hypersigma}, creating a direct trade-off between strict leakage-free evaluation and large-context architectures on a scene of this size.

\section{Related work}
\label{sec:related}

\subsection{Leakage in spatially autocorrelated data} 
The consequences of ignoring spatial dependence when partitioning data have been quantified most thoroughly in ecology and geostatistics. Roberts \etal~\cite{roberts2017crossvalidation} review cross-validation for dependent data and show that random folds seriously underestimate predictive error whenever a dependence structure is present, recommending spatial block designs in their place. Blocking alone is not sufficient, since samples on either side of a block boundary remain adjacent. The standard remedy is to pair blocking with a buffer that discards samples within a chosen distance of the boundary~\cite{valavi2019blockcv}. The buffer width, however, is specified as a distance between partitions rather than as the spatial extent a model actually consumes.
 
Within hyperspectral classification, Nalepa \etal~\cite{nalepa2019validating} showed that the repeated random splits commonly used in the literature can cause leakage between training and test sets. They proposed a patch-based partitioning algorithm; released ready-to-use folds for Salinas, Pavia University, and Indian Pines with a reference implementation; and evaluated a spectral and a spectral-spatial network on those folds. Zou \etal~\cite{zou2020ss3fcn} assigned spatial blocks to interleaved folds, extracted patches within blocks so that training and test patches do not overlap, and compared SS3FCN against several baselines under their own split. Both designs guarantee that training and test patches do not overlap. Still, neither separates the partitions themselves: a test pixel at a patch or block boundary can remain immediately adjacent to a training pixel. Neither study reports partition-level diagnostics for the separation it achieves, and neither asks whether the rankings reported in this literature change once leakage is removed.

\subsection{Architectures for HSI classification}
Support vector machines and random forests remain competitive baselines for HSI classification~\cite{melgani2004svm,belgiu2016random}. Spectral and spectral-spatial convolutional neural networks (CNNs) introduced learned feature hierarchies over the spectral axis and local
neighborhoods~\cite{roy2020hybridsn,zou2020ss3fcn}. Transformers adapted to the spectral sequence~\cite{hong2022spectralformer}, generic vision backbones were carried over from natural images~\cite{liu2022convnext}, state-space models were introduced for their linear-time treatment of long spectral sequences~\cite{wang2025s2mamba}, and large pretrained models have recently been proposed for hyperspectral interpretation~\cite{wang2025hypersigma}.
 
\subsection{Evaluation reliability in vision and machine learning}
Elsewhere in the field, revisiting evaluation has changed what the evidence supports. Gulrajani and Lopez Paz~\cite{gulrajani2021search} re-ran domain-generalization algorithms under matched conditions and found that none improved on empirical risk minimization by more than one percentage point of average accuracy, concluding that model selection is itself part of the problem. Barz and Denzler~\cite{barz2020cifair} found that $3.3\%$ of CIFAR-10 and $10\%$ of CIFAR-100 test images duplicate a training image, and released corrected test sets. In both cases, the contribution was not a new method but a demonstration that reported conclusions depended on the protocol.

Leakage-aware splits for hyperspectral classification exist and have been released~\cite{nalepa2019validating,zou2020ss3fcn}. Both guarantee that training and test patches do not overlap, but neither reports what spatial separation its partitions achieve. Nalepa \etal\ retain test pixels whose neighborhoods extend beyond a patch and zero-pad the missing neighbors, so boundary pixels are scored on incomplete input rather than excluded. Both splits were used to evaluate a small number of models, selected to validate each paper's own contribution.
\section{Methodology}
\label{sec:method}
 
Experiments are conducted on the corrected Salinas AVIRIS scene from the GIC Hyperspectral Remote Sensing Scenes collection, using the Zenodo-distributed copy~\cite{salinas_zenodo}. The scene comprises $512 \times 217$ pixels at $3.7$\,m ground sampling distance, with $16$ land-cover classes over $54{,}129$ labeled pixels. The corrected distribution retains $204$ of the $224$ acquired bands; the water absorption bands at one-indexed positions $108$--$112$, $154$--$167$, and $224$ have been removed. No header file ships with this distribution, so the wavelength axis is reconstructed as
\texttt{linspace(400, 2500, 224)} mapped onto the retained bands, with a grid uncertainty of roughly $\pm 20$\, nm; any figure carrying a wavelength axis inherits it.

\subsection{A leakage-free spatial partition}
\label{sec:partition}
 
Random pixel splitting assumes that a test pixel drawn from the same scene as the training pixels constitutes an independent observation. On an
agricultural scene at meter-scale resolution, this assumption fails twice over. Neighboring pixels of a single field are near-duplicates of one
another, so a test pixel sampled beside a training pixel supplies little evidence about generalization. Patch-based classifiers make the failure
concrete rather than statistical: a model consuming a $k \times k$ neighborhood around a test pixel physically reads training pixels during
inference, and has read the corresponding test neighborhood during fitting.
 
Protocol~B partitions the labeled pixels into class-aware contiguous chunks under seed $37$, separated by a per-class buffer that is assigned to
none of the training, validation, or test sets. The resulting partition contains $4{,}465$ training, $4{,}425$ validation and $41{,}542$ test pixels, with
$3{,}697$ pixels held in the buffer. Protocol~A is the random contrast, generated at seed $1037$ over the same labeled pixels, with raw partition sizes matched exactly to Protocol~B before admissibility masking.

\begin{figure}[t]
  \centering

  \begin{subfigure}{\linewidth}
    \centering
    \includegraphics[width=0.62\linewidth]{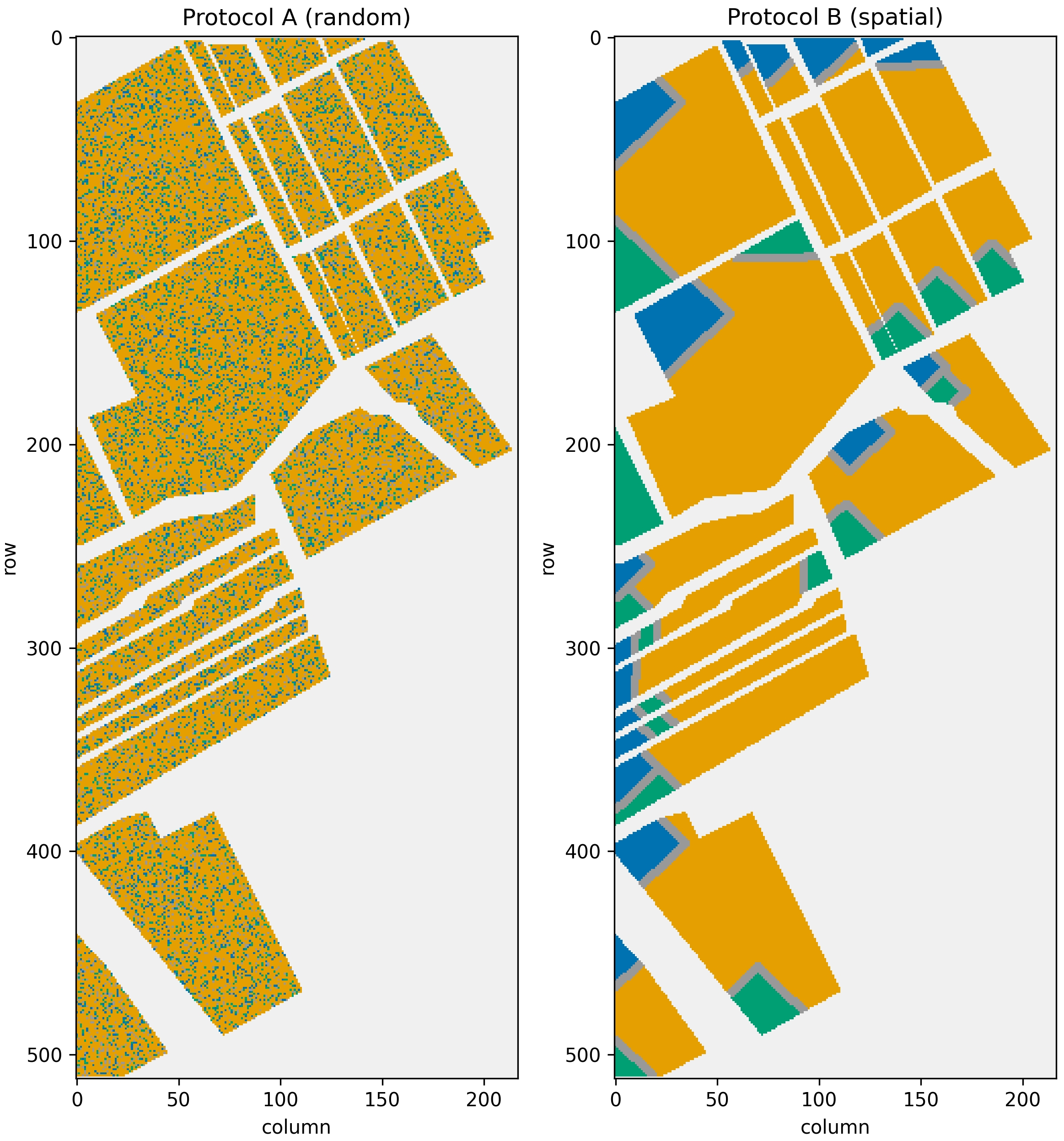}
    \caption{}
    \label{fig:protocol-map}
  \end{subfigure}

  \vspace{1mm}

  \begin{subfigure}{\linewidth}
    \centering
    \includegraphics[width=0.72\linewidth]{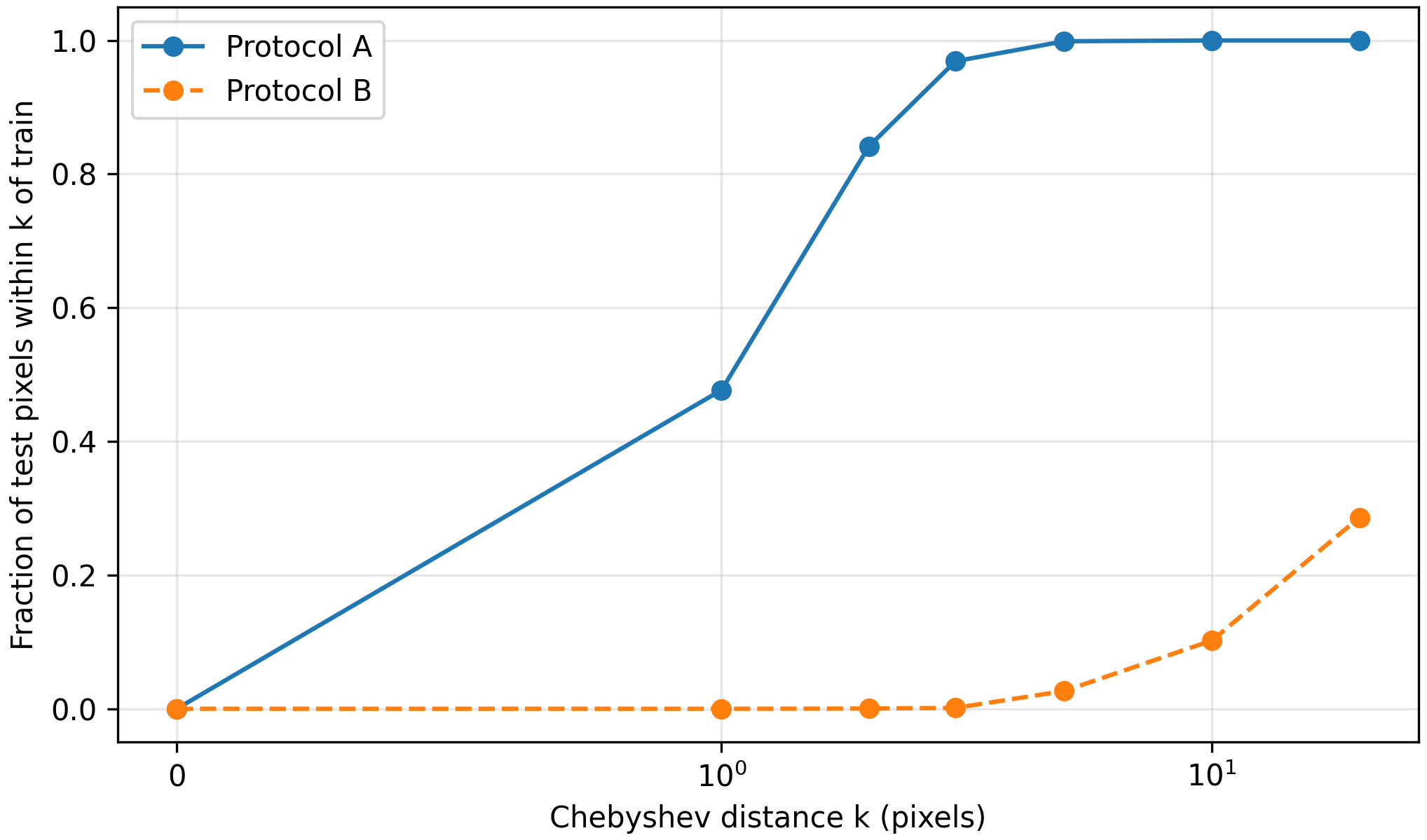}
    \caption{}
    \label{fig:protocol-curve}
  \end{subfigure}

  \caption{Protocol~A and Protocol~B on Salinas. (a) Partition assignment over the scene. Protocol~A distributes training
  and test pixels throughout, while Protocol~B assigns class-aware contiguous chunks separated by a per-class buffer.
  (b) Fraction of test pixels lying within Chebyshev distance $k$ of any training pixel. At $k=1$, $47.69\%$ of Protocol~A test pixels are
  immediately adjacent to a training pixel, compared with $0.00\%$ under Protocol~B. Raw partition sizes are matched between protocols before admissibility masking.}
  \label{fig:protocol}
\end{figure}

Disjointness is measured rather than assumed. The minimum Chebyshev distance between any training and any test pixel is $2$ for Protocol~B and $1$ for Protocol~A, the latter being the floor for disjoint pixel sets. \Cref{fig:protocol} reports the resulting contamination curves: at Chebyshev distance $1$, $47.69\%$ of Protocol~A test pixels are immediately adjacent to a training pixel, against $0.00\%$ under Protocol~B. The consequence is visible without training any model. A nearest-neighbor classifier on raw spectra, which has no spatial component whatsoever, scores $0.929$ Macro-F1 with Protocol~A and $0.819$ with Protocol~B. That $0.110$ difference is attributable to partitioning alone, showing that the protocol effect is visible before considering any architecture.

\subsection{Receptive-field admissibility}
\label{sec:admissibility}
 
Partition disjointness is necessary but not sufficient. Two pixels that belong to different partitions may still share input if the patches drawn
around them overlap, and whether they do is a function of the model's receptive field rather than of the partition. The criterion is elementary, but it is
not typically made explicit in hyperspectral evaluation protocols, where separation is specified independently of the models being evaluated:

\begin{quote}
Two pixels at Chebyshev distance $d$, each consumed as a patch of radius $r$, share at least one pixel if and only if $d \leq 2r$. Leakage-free evaluation therefore requires $d \geq 2r + 1$.
\end{quote}
 
A partition with minimum separation $2$ is thus leakage-free only for $r = 0$, which excludes every patch-based model in the benchmark. Rather than
discarding those models or relaxing the separation requirement, admissible pixels are selected explicitly. Under the
\texttt{own\_partition\_plus\_unlabeled} policy, a pixel is admissible at radius $R$ if its full $(2R+1) \times (2R+1)$ patch contains only pixels of
its own partition and unlabeled pixels, and no pixel of another partition or of the buffer. Unlabeled pixels are permitted because they carry no
class information and cannot transmit a label between partitions. The policy is monotone in $R$: a pixel admissible at radius $R$ is admissible at every
smaller radius, so a single subset constructed at the largest required radius serves every model in the benchmark.
 
The evaluation radius is fixed by a rule registered before any model was trained. Test retention must reach $50\%$ overall with at least $30$ pixels in every class, so that per-class estimates retain statistical power; training patch availability must reach $50\%$, so that models remain trainable; and validation must retain at least $10$ pixels per class, so that model selection has signal. The largest radius satisfying all three conditions is $R_\mathrm{eval} = 2$, yielding a common evaluation subset of $39{,}191$ test pixels at $94.34\%$ retention, with $69.49\%$ validation and $68.17\%$ training retention. Radius $3$ fails on validation class $11$, which retains no pixels at all.

\begin{figure}[t]
  \centering
  \includegraphics[width=0.72\linewidth]{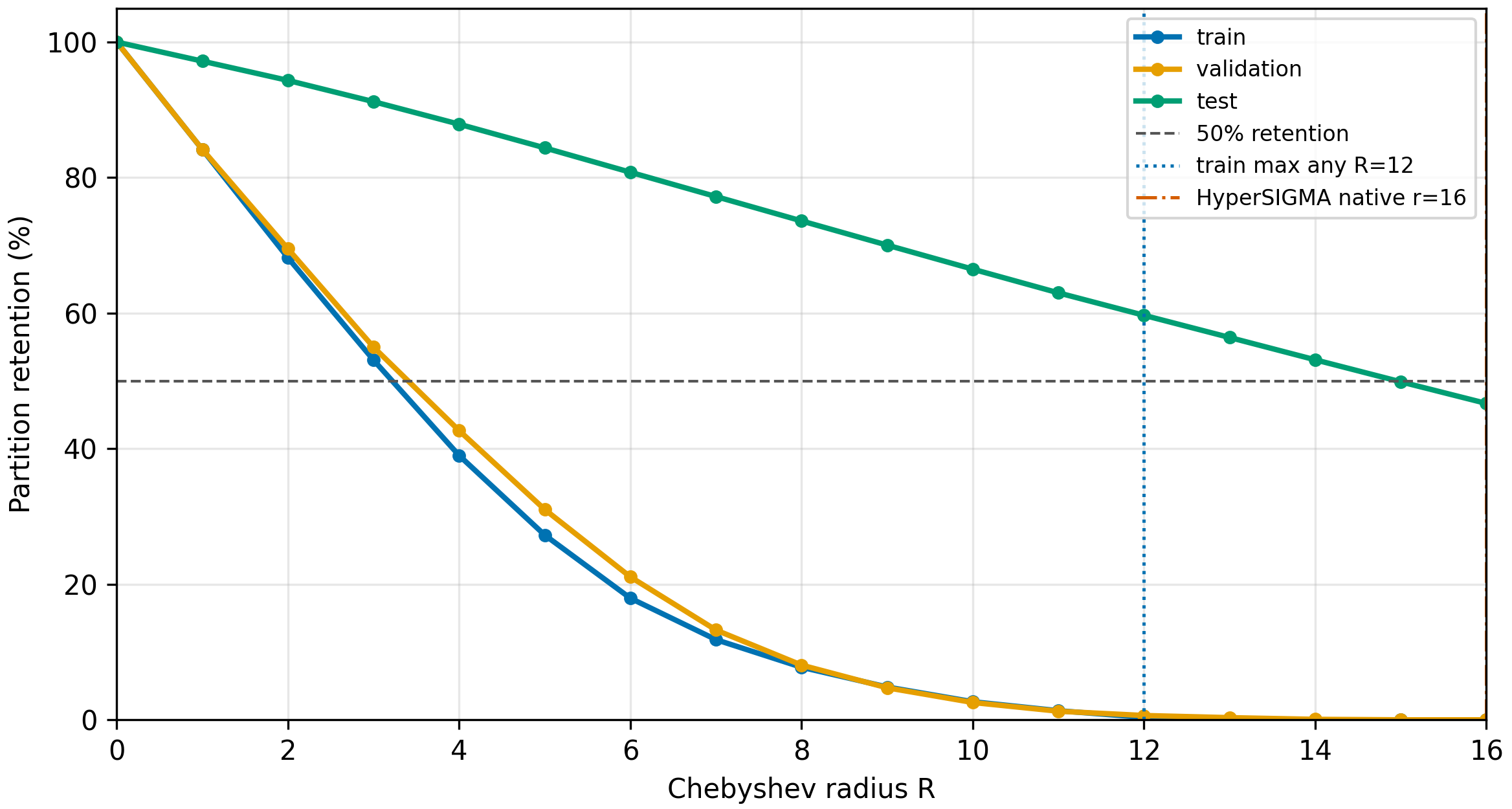}
  \caption{Patch availability by partition under the
  \texttt{own\_partition\_plus\_unlabeled} policy. Retention falls faster for the training and validation partitions than for the test partition, which occupies larger contiguous regions. The binding constraint on the evaluation radius therefore originates in training and model selection rather than in the test set: the training
partition retains at least half its pixels only to $R=3$, and validation class $11$ is empty at $R=3$, so the pre-registered rule fixes $R_\mathrm{eval}=2$. The native radius of the foundation model considered for the benchmark exceeds the training partition's support ceiling entirely.}
  \label{fig:radius}
\end{figure}

\Cref{fig:radius} shows why the binding constraint arises where it does. Patch availability decays far faster for training and validation than for test, because the test partition occupies larger contiguous regions while training and validation are distributed as narrower chunks. The training partition admits patches up to radius $12$ in principle, but retains at least half its pixels only to radius $3$. The constraint on this benchmark therefore originates in the training and model-selection partitions rather than in the test set, which has a consequence examined in \cref{sec:results}: architectures whose published configurations require large windows cannot be trained at their native radius on a leakage-free split of a scene this size, irrespective of how much test data survives.

Masking a benchmark invites the objection that the retained subset is easier. The direction was measured rather than argued. Of the pixels removed at $R_\mathrm{eval} = 2$, only $1.1\%$ lie on a cross-class boundary; the remainder are interior pixels adjacent to the training or buffer regions. Evaluated with the spectral probes, the removed pixels score approximately six Macro-F1 points \emph{higher} than the retained ones. Masking therefore removes easy pixels, and the common evaluation subset is a harder test set than the full partition, not a more permissive one.

\subsection{Benchmark protocol}
\label{sec:protocol}
 
Ten architectures are evaluated, spanning linear discriminant, kernel, and ensemble baselines; spectral networks; a spectral transformer; spectral-spatial convolutional networks; a generic vision backbone; and a state-space model (\cref{tab:roster}). Two radii are distinguished: $R_\mathrm{eval}$, the global admissibility radius used to build the common subset, fixed at $2$; and $r_\mathrm{eval} = \min(r_\mathrm{nat}, R_\mathrm{eval})$, the radius an individual model consumes. They differ only for the pixel-wise models, which are scored on the same masked subset so that under Protocol~B all ten are evaluated on identical pixels. Radii are determined from released sources: where an architecture accepts a fixed input window, that window bounds the radius regardless of internal depth, so HybridSN, which accumulates an internal radius of $4$ but consumes a $25 \times 25$ window, has effective radius $12$. Per-model radii and their source locations are released with the benchmark. Models fall into two tiers by the severity of the constraint at $R_\mathrm{eval} = 2$. Tier~1 contains the seven models evaluated at or within one pixel of their native radius. Tier~2 contains the three whose native radius exceeds $2$ by more than one pixel; their results are reported as evaluations at $5 \times 5$, not of the published configurations.

\begin{table}
  \centering
  \small
  \setlength{\tabcolsep}{4pt}

  \begin{tabular}{@{}llccrl@{}}
    \toprule
    Model & Family & $r_{\mathrm{nat}}$ & $r_{\mathrm{eval}}$ & Params & Prov. \\
    \midrule
    \multicolumn{6}{@{}l}{\emph{Tier~1: native or near-native}} \\
    PLS-DA~\cite{barker2003pls}         & linear disc.      & 0  & 0   & 3{,}520   & sk \\
    RBF-SVM~\cite{melgani2004svm}        & kernel            & 0  & 0   & 117{,}504 & sk \\
    RandomForest~\cite{belgiu2016random}   & ensemble          & 0  & 0   & 22{,}208  & sk \\
    CNN1D          & spectral          & 0  & 0   & 37{,}552  & base  \\
    ResNet1D~\cite{he2016deep}       & spectral          & 0  & 0   & 254{,}384 & port \\
    SS3FCN~\cite{zou2020ss3fcn}         & spec.-spat.       & 3  & 2   & 76{,}496  & port \\
    SpectralFormer~\cite{hong2022spectralformer} & spec. transformer & 3  & 2   & 215{,}952 & port \\
    \addlinespace

    \multicolumn{6}{@{}l}{\emph{Tier~2: severely constrained}} \\
    S$^2$Mamba~\cite{wang2025s2mamba}     & state-space       & 5  & 2   & 252{,}304 & ref \\
    ConvNeXt2D~\cite{liu2022convnext}     & vision backbone   & 7  & 2   & 123{,}536 & port \\
    HybridSN~\cite{roy2020hybridsn}       & spec.-spat.      & 12 & 2   & 49{,}024  & port \\
    \addlinespace

    \multicolumn{6}{@{}l}{\emph{Structurally excluded}} \\
    HyperSIGMA~\cite{wang2025hypersigma}     & foundation        & 16 & n/a & n/a       & n/a \\
    \bottomrule
  \end{tabular}

  \vspace{3pt}

  {\footnotesize
  \begin{tabular}{@{}ll@{}}
    $r_{\mathrm{nat}}$  & native radius: max of input window or layer-accumulated radius \\
    $r_{\mathrm{eval}}$ & evaluated radius, capped at $R_{\mathrm{eval}}=2$ \\
    sk                  & scikit-learn implementation \\
    base                & generic baseline \\
    port                & reimplemented from published description \\
    ref                 & official reference implementation \\
    n/a                 & omitted, no valid configuration at $R_{\mathrm{eval}}=2$ \\
  \end{tabular}
  }

  \caption{Benchmark models, ordered by native receptive radius.}
  \label{tab:roster}
\end{table}

\Cref{tab:roster} records how each model was obtained: three come from \textsf{scikit-learn}, five are reimplementations of published descriptions, released with the benchmark so that fidelity can be checked, and S$^2$Mamba runs on a reference selective-scan implementation rather than the fused CUDA kernel. CNN1D is a generic spectral baseline with no canonical source and is implemented directly. One model is excluded outright. HyperSIGMA, the pretrained hyperspectral model in the benchmark, admits no configuration of this split at $R_\mathrm{eval} = 2$: its published tokenization cannot accept a $5 \times 5$ window, and its native radius of $16$ exceeds the training partition's ceiling of $12$.

Each model receives exactly $90$ Optuna trials per table, using a Tree-structured Parzen Estimator (TPE) sampler at a fixed seed, with median pruning applied to neural models and pruned-trial counts reported. Convergence is verified rather than assumed:
The index of the best trial is reported for every model, and any model whose optimum falls in the final decile of the search is flagged. The budget was
set at $90$ after a $30$-trial pilot proved equal across models but did not converge, and produced a ranking that changed once the budget was raised.

Model selection and early stopping use validation data only; no test score enters the selection path. Preprocessing statistics are fit on training pixels only, and the choice of preprocessing is placed inside the search space rather than fixed per model, so no architecture benefits from a hand-selected transform. Test scores are recorded for every trial but read only after the search concludes, and are used solely to report the selected configuration and to compute the validation-test correlation as a diagnostic for selection overfitting.

Four evaluations are reported. The primary one scores every model on the Protocol~B common subset and is the only head-to-head comparison. A second scores the $r_\mathrm{eval}=0$ models on the full unmasked Protocol~B partition, isolating the cost of masking. A third repeats the primary evaluation under Protocol~A at the same $R_\mathrm{eval}=2$, so the training partitions differ only in protocol. Protocol~B scores the $39{,}191$-pixel common subset of pixels, while Protocol~A is unmasked and scores $41{,}542$ pixels for pixel-wise models and $40{,}948$ for patch-based models after border filtering; the two sets overlap on $30{,}057$ pixels. Masking also removes roughly a third of the training data. Hence, a fourth repeats the Protocol~A evaluation with training and validation subsampled to the counts Protocol~B leaves available ($3{,}044$ and $3{,}075$), stratified by class and resampled per seed. Protocol~A at matched training size is the fair contrast; the difference between the two Protocol~A evaluations isolates the size effect.

Macro-F1 is the primary metric, since class support on Salinas ranges from $916$ to $11{,}271$ pixels and overall accuracy is dominated by the largest classes. Overall accuracy, average accuracy, Cohen's $\kappa$, and expected calibration error are reported alongside it; the last is omitted for PLS-DA, whose softmax over decision scores is not a probability; per-class and per-metric breakdowns for all four evaluations are released with the benchmark. Every cell runs over five seeds ($37$, $137$, $237$, $337$, $437$), reported as mean and standard deviation. Models are compared pairwise with McNemar's test on per-pixel predictions under Holm correction, reported as odds ratios: across $39{,}191$ predictions, small differences are significant but not meaningful. Bootstrap intervals over test pixels are reported separately from seed variance, since the two measure different things.

The frozen partition, the hash-verified common evaluation subset, the pre-registered selection rules, an environment lockfile, and the entry-point script with fixed seeds are released with the paper; the supplementary material lists contents and hashes.
\section{Results}
\label{sec:results}
 
Three settings are compared in \cref{tab:main}. Protocol~B is the leakage-free evaluation on the common subset from \cref{sec:admissibility} and is the only setting in which all ten models are scored on identical pixels. Protocol~A at full training size provides the literature-style reference ranking. Protocol~A at matched training size repeats
that evaluation at the same $R_\mathrm{eval}=2$, with training and validation reduced to the counts available under Protocol~B, so it controls for training-set size when comparing the two protocols. Rank comparisons use Protocol~A at full training size, while performance gaps use Protocol~A at matched training size (\cref{tab:main}). A fourth setting, unmasked Protocol~B, applies only to the models that consume no spatial context and is used once to measure the cost of masking.
 
\subsection{The cost of leakage-free evaluation}
Every architecture scores lower under Protocol~B than under Protocol~A
(\cref{tab:main}). The mean drop is $0.165$ Macro-F1 at full training size and $0.147$ at matched size; the $0.018$ difference is the cost of the smaller training set. Masking might be expected to make the subset easier; it does the opposite. The mask can be lifted only for the five $r_\mathrm{eval}=0$ models, since a patch-based model scored unmasked would read training pixels through its receptive field. All five improve by $0.003$ to $0.091$ Macro-F1, consistent with the spectral analysis in \cref{sec:admissibility}. The loss also varies by model: PLS-DA drops only $0.074$, while SpectralFormer and HybridSN drop $0.223$ and $0.228$, enough to change the ranking.

\begin{table*}[t]
  \centering
  \small
  \setlength{\tabcolsep}{5pt}
  \begin{tabular}{@{}llccccccc@{}}
    \toprule
    & & \multicolumn{2}{c}{Protocol~B} & \multicolumn{2}{c}{Protocol~A} & &
    \multicolumn{2}{c}{Matched A} \\
    \cmidrule(lr){3-4}\cmidrule(lr){5-6}\cmidrule(lr){8-9}
    Model & Tier & Macro-F1 & rank & Macro-F1 & rank & $\Delta$ & Macro-F1 & Gap \\
    \midrule
    ConvNeXt2D      & 2 & $0.848 \pm 0.013$ & 1  & $0.988 \pm 0.002$ & 1  & $0$  & $0.984 \pm 0.002$ & $0.136$ \\
    S$^2$Mamba      & 2 & $0.847 \pm 0.007$ & 2  & $0.983 \pm 0.001$ & 2  & $0$  & $0.979 \pm 0.002$ & $0.132$ \\
    ResNet1D        & 1 & $0.837 \pm 0.008$ & 3  & $0.958 \pm 0.002$ & 5  & $+2$ & $0.945 \pm 0.003$ & $0.109$ \\
    RBF-SVM         & 1 & $0.813 \pm 0.000$ & 4  & $0.967 \pm 0.000$ & 4  & $0$  & $0.960 \pm 0.002$ & $0.147$ \\
    RandomForest    & 1 & $0.796 \pm 0.005$ & 5  & $0.954 \pm 0.000$ & 6  & $+1$ & $0.947 \pm 0.001$ & $0.151$ \\
    CNN1D           & 1 & $0.791 \pm 0.034$ & 6  & $0.945 \pm 0.001$ & 7  & $+1$ & $0.916 \pm 0.012$ & $0.126$ \\
    SS3FCN          & 1 & $0.750 \pm 0.023$ & 7  & $0.941 \pm 0.003$ & 9  & $+2$ & $0.898 \pm 0.019$ & $0.148$ \\
    SpectralFormer  & 1 & $0.732 \pm 0.013$ & 8  & $0.971 \pm 0.001$ & 3  & $\mathbf{-5}$ & $0.955 \pm 0.006$ & $0.223$ \\
    HybridSN        & 2 & $0.703 \pm 0.054$ & 9  & $0.944 \pm 0.008$ & 8  & $-1$ & $0.930 \pm 0.018$ & $0.228$ \\
    PLS-DA          & 1 & $0.675 \pm 0.000$ & 10 & $0.793 \pm 0.000$ & 10 & $0$  & $0.749 \pm 0.004$ & $0.074$ \\
    \midrule
    Benchmark mean     &   & $0.779$ &  & $0.944$ &  &  & $0.926$ & $0.147$ \\
    \bottomrule
  \end{tabular}
  \caption{Main results, ordered by Protocol~B Macro-F1 (mean $\pm$ std over five seeds). Protocol~A is evaluated at full training size, the setting commonly reported in the literature; Matched~A repeats it with training and validation subsampled to the counts Protocol~B leaves available. $\Delta$ is the rank change from Protocol~A to Protocol~B, positive where a model improves under leakage-free evaluation, and Gap is the loss against Matched~A. Protocol~A is unmasked, so only Protocol~B scores every model on identical pixels (\cref{sec:protocol}). The mean drop is $0.165$ against Protocol~A and $0.147$ against Matched~A, the difference being the cost of the smaller training set. Spearman correlation between the Protocol~B and Protocol~A orderings is $0.78$. Tier follows \cref{tab:roster}; zero deviation reflects deterministic fitting.}
  \label{tab:main}
\end{table*}

\subsection{Rankings do not survive the correction}
Because the losses differ across models, the ranking changes. The Spearman
correlation between Protocol~A (full training size) and Protocol~B is $0.78$, with shifts of up to five places: SpectralFormer falls from third to eighth, while ResNet1D moves from fifth to third. The matched comparison shows the same pattern ($\rho=0.77$), so the smaller training set does not cause the change. A tuned RBF-SVM ranks fourth under Protocol~B, ahead of six of the nine other models, including two spectral-spatial CNNs and a spectral transformer. It ranks fourth under Protocol~A as well, but there the six it leads trail it by margins an order of magnitude smaller.

To test whether the limited $5\times5$ context explains these results,
ResNet1D, SpectralFormer, and SS3FCN were evaluated at mask radii $R=0$, $1$, and $2$. Neither patch-based model improves with larger context: SpectralFormer drops by $0.13$ and SS3FCN by $0.08$ Macro-F1, while ResNet1D remains stable. For these models, limited spatial context therefore does not explain their position in \cref{tab:main}; this includes SpectralFormer, whose five-place fall is the largest in the benchmark. The severely constrained Tier~2 models were not included in this sweep. Full curves are provided in the supplementary material.

\subsection{Architectures diverge, yet fail on the same pixels}
Architecture matters under Protocol~B: the best and worst models differ by
$0.173$ Macro-F1, against only $0.0157$ average variation across seeds. The
differences in \cref{tab:main} are eleven times larger than seed noise, meaning the performance gaps mainly reflect architectural differences rather than random variation.

The errors are also strongly shared. The mean pairwise overlap of misclassified pixels is $0.471$, increasing to $0.537$ among the four strongest models, which span kernel, spectral, state-space, and vision-backbone families. Of all pixels misclassified by at least one model, $30\%$ are missed by seven or more.

\begin{figure}[t]
  \centering
  \includegraphics[width=\linewidth]{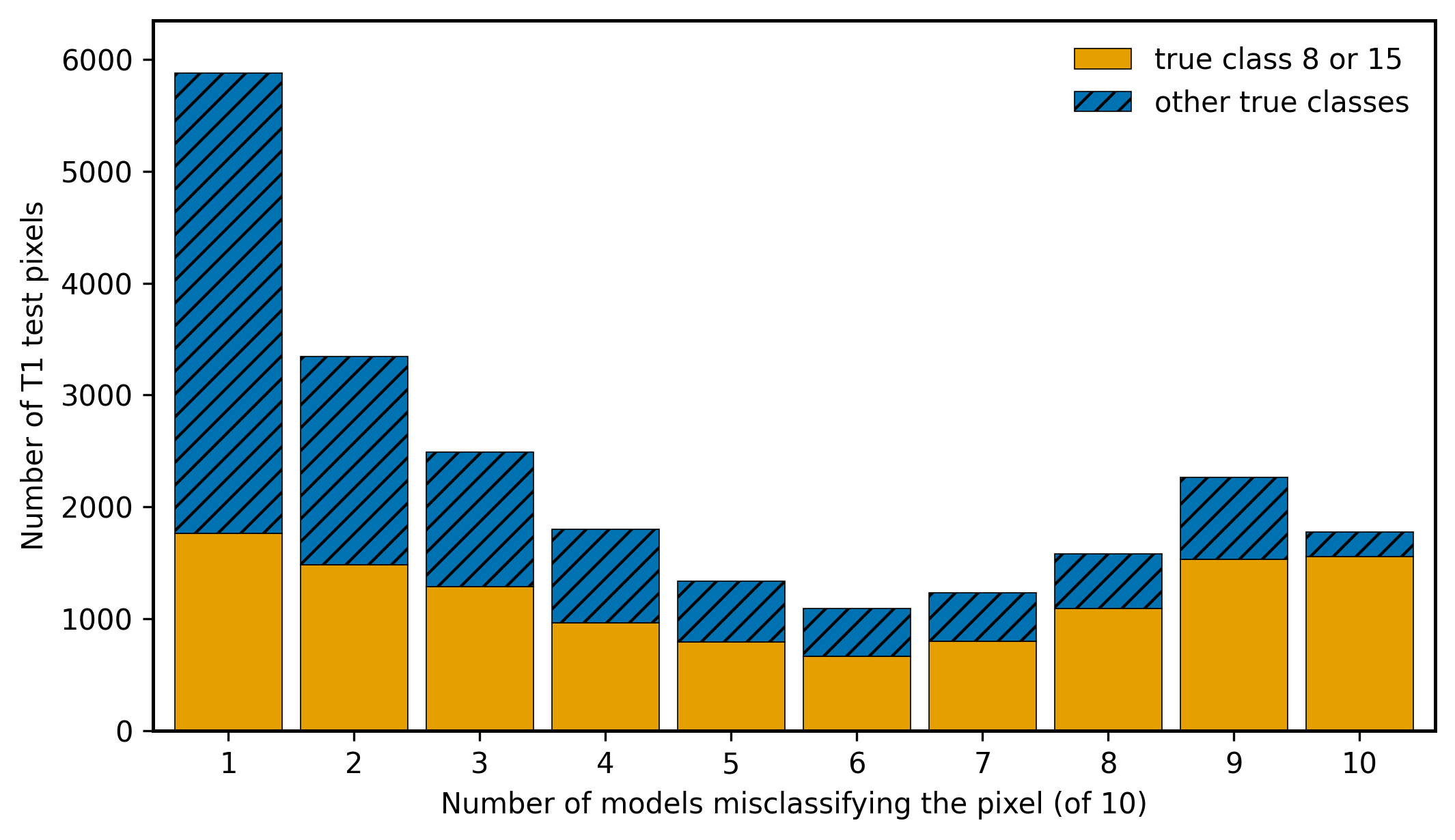}
  \caption{Number of models misclassifying each Protocol~B test pixel, with the contribution of classes~8 and~15 shown separately. The leftmost bar contains the $16{,}394$ pixels that no model misclassifies. Over the remaining pixels, the distribution is bimodal: counts fall to a minimum at six models and rise again at nine and ten, marking a core of $1{,}779$ pixels misclassified by all
ten models. Seed~37.}
  \label{fig:agreement}
\end{figure}

\Cref{fig:agreement} shows a clear core of hard pixels: error counts fall until six models, then rise again at nine and ten. The classes involved change along the way. Where only one model fails, the errors are spread across the scene; where all ten fail, almost every pixel belongs to class~8 or class~15. On seed~37, $1{,}779$ pixels are missed by all ten models, giving $17{,}790$ wrong predictions. Of these, $83\%$ are class~8 predicted as class~15, and the confusion runs one way: $14{,}842$ such predictions against only $424$ in reverse. The affected pixels are $17.6\%$ of all class-8 test pixels.
 
The two classes are never spatially adjacent. Their minimum Chebyshev distance is $3$, so no model in the benchmark sees them in the same patch, and the confusion is spectral rather than spatial. The pattern is stable across seeds: of the $1{,}779$ pixels missed by every model on seed~37, $1{,}156$ are missed by every model on all five.

\subsection{The protocol bounds receptive field}
The limit noted in \cref{sec:admissibility} appears in the benchmark models.
The training, validation, and test partitions support maximum radii of $12$,
$14$, and $16$ pixels, while the training set retains half its pixels only up
to radius $3$. HyperSIGMA requires radius $16$, so the training split cannot
support it, and its published tokenization also cannot use the $5\times5$
window allowed by $R_\mathrm{eval}=2$. This split therefore cannot evaluate
HyperSIGMA.

The limit comes from the partition geometry, not the model. Any spatially
disjoint split restricts the receptive field available for training, and the
restriction tightens as the scene gets smaller. On Salinas it already falls
below what a large-context foundation model requires, placing leakage-free
evaluation and such architectures in tension at this scene size.
 
\subsection{Supporting analyses}
Of the $45$ pairwise comparisons on the Protocol~B common subset, $44$ are
significant under McNemar's test with Holm correction; only HybridSN and SS3FCN are indistinguishable ($\chi^2 = 0.96$, $p_\mathrm{Holm} = 0.33$). Significance at this sample size says little about practical difference. ConvNeXt2D and S$^2$Mamba separate at $p_\mathrm{Holm} \approx 10^{-140}$ with an odds ratio of $2.30$ over $4{,}142$ discordant pixels, yet differ by $0.0007$ Macro-F1 on the seed mean, well within the seed variation of either model.

Neither capacity nor calibration predicts accuracy under Protocol~B. ConvNeXt2D and S$^2$Mamba, the two strongest models, differ by a factor of two in parameters, and the largest model in the benchmark is neither the most nor the least accurate. Expected calibration error ranges from $0.022$ to $0.234$ among the neural models, and the best-calibrated model places fifth on Macro-F1.
\section{Discussion}
\label{sec:discussion}
 
Taken together, the results reveal two sources of difficulty hidden by random splitting. The first comes from the evaluation protocol itself. At matched training size, the ten models score $0.147$ Macro-F1 lower under Protocol~B than under Protocol~A. The same pattern appears even without a trained model:
The nearest-neighbor classifier on raw spectra, which uses no spatial information at all, scores $0.110$ higher under random splitting than under spatially disjoint evaluation. Part of the performance reported on Salinas using random splitting therefore comes from the split itself, not only from what the models learn.
 
The second source is the data itself. Architecture choice still matters under Protocol~B. The performance spread across models is eleven times larger than the across-seed variation. Yet the models fail on largely the
same pixels, and $1{,}156$ pixels are misclassified by every architecture across all seeds. Classes~8 and~15 are never spatially adjacent, their minimum separation being $3$ pixels, so no model in the benchmark sees them in the same patch. The shared error set therefore points to spectral ambiguity rather than a partition artifact. Different architectures shift errors around this difficult subset, but none resolves it.
 
Two limitations define the scope of these results. First, the experiments use one scene and one frozen partition, so the reported effect sizes are specific to Salinas, while the admissibility criterion and audit procedure are general. Second, the two highest-ranked models, ConvNeXt2D and S$^2$Mamba, are evaluated with $5\times5$ inputs rather than their published $15\times15$ and $11\times11$ windows. Their results therefore reflect the architectures under a receptive-field constraint. The radius-sensitivity test in \cref{sec:results} found no gain from additional context for the models it covered, but neither of these two was among them.

Two reporting practices follow. First, spatial separation should be reported together with model's receptive field. A split with minimum separation $d$ is leakage-free only for models whose evaluated patch radius satisfies $r_\mathrm{eval} \leq (d-1)/2$; Protocol~B satisfies this per pixel through the mask of \cref{sec:admissibility} rather than globally. One quantity without the other leaves a protocol incomplete.

Second, leakage-free evaluation sets a measurable limit on the architectures a benchmark can support, and that limit is geometric rather than a consequence of any particular design. Guaranteeing a model a clean patch means discarding every pixel near the edge of its partition region, and the strip lost this way widens with patch size. Once the patch approaches the width of the regions themselves, almost nothing survives. Changing the train/validation/test proportions widens some regions and narrows others: it shifts which partition runs out first and the radius at which this happens, but it cannot remove that limit. Under Protocol~B, the training partition runs out first, keeping half its pixels only up to radius $3$ and admitting no patch at all beyond radius $12$. Both figures follow from the split geometry and can be computed before choosing any model. Reporting them shows which architectures a benchmark can host and which require a different split or a larger scene.
\section{Conclusion}
\label{sec:conclusion}
This work examined whether the main conclusions drawn from hyperspectral classification on Salinas hold with leakage-free spatial evaluation. The results show a clear shift. In the random split constructed here, $47.69\%$ of test pixels lie within one pixel of a training pixel; under the leakage-free partition, none do. Removing this adjacency lowers Macro-F1 by $0.110$ for a nearest-neighbor classifier before model architecture enters the comparison, and by $0.147$ on average across the ten models at matched training size. Model rankings shift by as many as five places, and a tuned RBF-SVM ranks above six of the nine other models in the benchmark. Performance on Salinas depends strongly on the evaluation protocol, and a single accuracy figure does not convey that.

The models also fail on many of the same pixels. Across all seeds, every architecture misclassifies $1{,}156$ pixels. Most come from one directional confusion between two classes that are not spatially adjacent within any model's receptive field, pointing to a shared spectral ambiguity rather than an artifact of the split. None of the ten architectures resolves it.

The exact accuracy drops are specific to Salinas, but the underlying issue is broader. Random splitting places correlated test and training pixels close together in any spatially structured scene, and any leakage-free split of such a scene has a receptive-field ceiling that no choice of partition proportions removes. The audit used here applies directly to other benchmarks, requiring only the distances in the split and the patch size each model reads. The released protocol, audit procedure, and common evaluation subset also allow future models to be compared on the same pixels.
{
    \small
    \bibliographystyle{ieeenat_fullname}
    \bibliography{main}
}

\end{document}